\documentclass[runningheads]{llncs}
\usepackage{amsmath,amssymb,amsfonts}
\usepackage{algorithmic}

\usepackage{textcomp}
\usepackage{xcolor}

\usepackage[T1]{fontenc}
\usepackage[utf8]{inputenc}
\usepackage[english]{babel}
\usepackage[bookmarks=false]{hyperref}       
\usepackage{url}            
\usepackage{booktabs}       
\usepackage{amsfonts}       
\usepackage{nicefrac}       
\usepackage{microtype}      
\usepackage{multicol}
\usepackage{lipsum}
\usepackage{csquotes}
\usepackage{graphicx}
\usepackage{caption}
\usepackage{subcaption}
\usepackage{multirow}
\usepackage{amssymb}
\usepackage[export]{adjustbox}

\usepackage{pifont}
\newcommand{\xmark}{\ding{55}}

\newcommand\modelname{Simple Predict \& Align Network}
\newcommand\modelacc{SPAN}
\newcommand\tens[1]{#1}

\begin{document}

\title{\modelacc{}: a \modelname{} for Handwritten Paragraph Recognition}

\author{Denis Coquenet\inst{1,2,3}\orcidID{0000-0001-5203-9423} \and
Clément Chatelain\inst{1,4}\orcidID{0000-0001-8377-0630} \and
Thierry Paquet\inst{1,2}\orcidID{0000-0002-2044-7542}}

\institute{LITIS Laboratory - EA 4108, France \and Rouen University, France \and Normandie University, France \and INSA of Rouen, France\\
    \email{\{denis.coquenet, clement.chatelain, thierry.paquet\}@litislab.eu}
}

\maketitle
\thispagestyle{plain}
\pagestyle{plain}

\begin{abstract}

Unconstrained handwriting recognition is an essential task in document analysis. It is usually carried out in two steps. First, the document is segmented into text lines. Second, an Optical Character Recognition model is applied on these line images. We propose the \modelname{}: an end-to-end recurrence-free Fully Convolutional Network performing OCR at paragraph level without any prior segmentation stage. The framework is as simple as the one used for the recognition of isolated lines and we achieve competitive results on three popular datasets: RIMES, IAM and READ 2016. The proposed model does not require any dataset adaptation, it can be trained from scratch, without segmentation labels, and it does not require line breaks in the transcription labels. Our code and trained model weights are available at \url{https://github.com/FactoDeepLearning/SPAN}.
\end{abstract}

\section{Introduction}
Offline handwritten text recognition consists in recognizing the text from a scanned document.
This task is usually performed in two steps, by two different neural networks. In a first step, the document image is cut into text regions: this is the segmentation step. Then, Optical Character Recognition (OCR) is applied on each text region images. Following the advances of the recognition process over time, segmentation was performed on larger and larger entities, from the character in the early ages, to text lines more recently, gradually decreasing the amount of segmentation labels required to train the system.

As a matter of fact, producing segmentation labels by hand, in addition to transcription labels, is costly. Moreover, the use of a two-step process requires to clearly define what a line should be in a non-latent pivot format, \textit{i.e}. a line of text, to generate target labels. However, the definition of a text line raises several questions that prevent to optimize its detection in order to maximize the recognition performance: is a text line a bounding box, a polygon, a set of pixels or a baseline? How should it be measured? Which loss should it be trained with? Given all these open questions, we claim that segmentation and recognition should be trained in an end-to-end fashion using a latent space between both stages. Indeed, this allows to circumvent any text line definition, while leveraging the annotation needs.

In this paper, we propose the \modelname{} (\modelacc{}), an end-to-end recurrence-free Fully Convolutional Network (FCN) model free of those issues, further reducing the needs for labels in two ways. First, the proposed model performs OCR at paragraph level, so it does not need line-level segmentation labels. Second, it does not even require line breaks in the transcription labels. The proposed model totally circumvents the line segmentation problem using a very straightforward approach. The input paragraph image is analysed in a 2D fashion using a classical fully convolutional architecture, leading to a 2D latent space that is reshaped into a 1D sequential latent space. Finally, the CTC loss is simply used to align the 1D character prediction sequence with the paragraph transcription.

This paper is organized as follows. Related Works are presented in Section \ref{related_work}. The proposed \modelacc{} architecture is described in Section \ref{architecture}. Section \ref{experiments} presents the experimental environment and the results. We draw conclusion in Section \ref{conclusion}.
\section{Related Works}
\label{related_work}
In the literature, multi-line text recognition is mainly carried out in two steps. First, a text region (line/word) segmentation is performed \cite{Renton2018,DHSegment,ARUNet}; then, an OCR is applied on the extracted text regions images \cite{Coquenet2019,Coquenet2020,Yousef_line,Michael2019} thanks to the Connectionist Temporal Classification (CTC) \cite{CTC}. As shown in \cite{Coquenet2021}, deep neural networks perform well on both task separately but, when put together, errors in the segmentation stage leads to errors in the OCR stage, leading to higher Character Error Rate (CER).

Recently, one can notice a trend towards the use of unified models. We can classify them into two categories: those performing a text region segmentation prior to the recognition and those without explicit segmentation.

\subsection{Segmentation-based approaches}
Segmentation-based approaches, by definition, require line or word segmentation labels, in addition to the associated transcription label; so the line breaks must be annotated. 

Among these approaches, \cite{Carbonell2019,Carbonell2020,Chung2020} are based on object-detection methods: a Region Proposal Network (RPN), followed by a non-maximal suppression process and Region Of Interest (ROI), generates line or word bounding boxes. An OCR is then applied on these bounding boxes.

Other approaches are based on predicting the start-of-line coordinates. While in \cite{Moysset2017} the line is considered horizontal, in \cite{Wigington2018,Wigington2019}, lines are normalized, recurrently predicting coordinates. Finally, an OCR is applied on these lines.

\subsection{Segmentation-free approaches}
Since they do not explicitly segment the input image, segmentation-free approaches do not require any segmentation labels; the models can be trained using transcription labels only.

In \cite{Bluche2016,Coquenet2021}, the proposed models incorporate an attention mechanism to recurrently generate line features, performing a kind of implicit line segmentation. Indeed, an encoder generates features from the input image; then, the attention process sequentially selects features to focus on. Finally, a decoder predicts characters from these features. \cite{Bluche2017} proposed a similar approach with an implicit character segmentation.

To our knowledge, only two other works have proposed segmentation-free approaches for multi-line text recognition. \cite{Schall2018} focuses on the loss to tackle the two-dimensional aspect of the task, providing a Multi-Dimensional Connectionist Classification (MDCC). Ground truth transcription labels are converted to a two-dimensional model, using a Conditional Random Field (CRF). It enables to jump from one line to the next one, adding a new line separator label in addition to the standard CTC blank label. In \cite{Yousef2020}, the model is trained to unfold the input multi-line text image into a sequence of lines, thus forming a single large line. Thus, the task is reduced to a one-dimensional problem and the model can be trained with the standard CTC loss.

\cite{Schall2018} and \cite{Bluche2017} are part of the first works proposed for multi-line text recognition, but they remain below the state of the art. While \cite{Bluche2016} requires pretraining on line-level images, \cite{Coquenet2020} requires line breaks in the transcription labels. \cite{Yousef2020} is the only model that can be trained from scratch, without any segmentation labels nor line breaks in the transcription labels; but, as a counterpart, it requires some hyperparameters to be adapted for each dataset. Indeed, it requires input images of fixed sizes and includes intermediate bilinear interpolations with fixed dimensions which are specific to each dataset.

In this work, we propose an end-to-end model trained in the same conditions, \textit{i.e.} with paragraph transcription as the only used label, without any line breaks. Instead of unfolding the input image as in \cite{Yousef2020}, we propose to train a model to both predict and align characters so as to get vertical separation between lines, preserving the two-dimensional nature of the task. Contrary to the work presented in \cite{Yousef2020}, the proposed model is able to handle variable size input images, making it flexible enough to be used on multiple datasets without modifying any hyperparameter.
\section{SPAN Architecture}
\label{architecture}

\begin{figure*}[htbp!]
\centering
    \begin{subfigure}[b]{\textwidth}
    \centering
    \includegraphics[width=\textwidth]{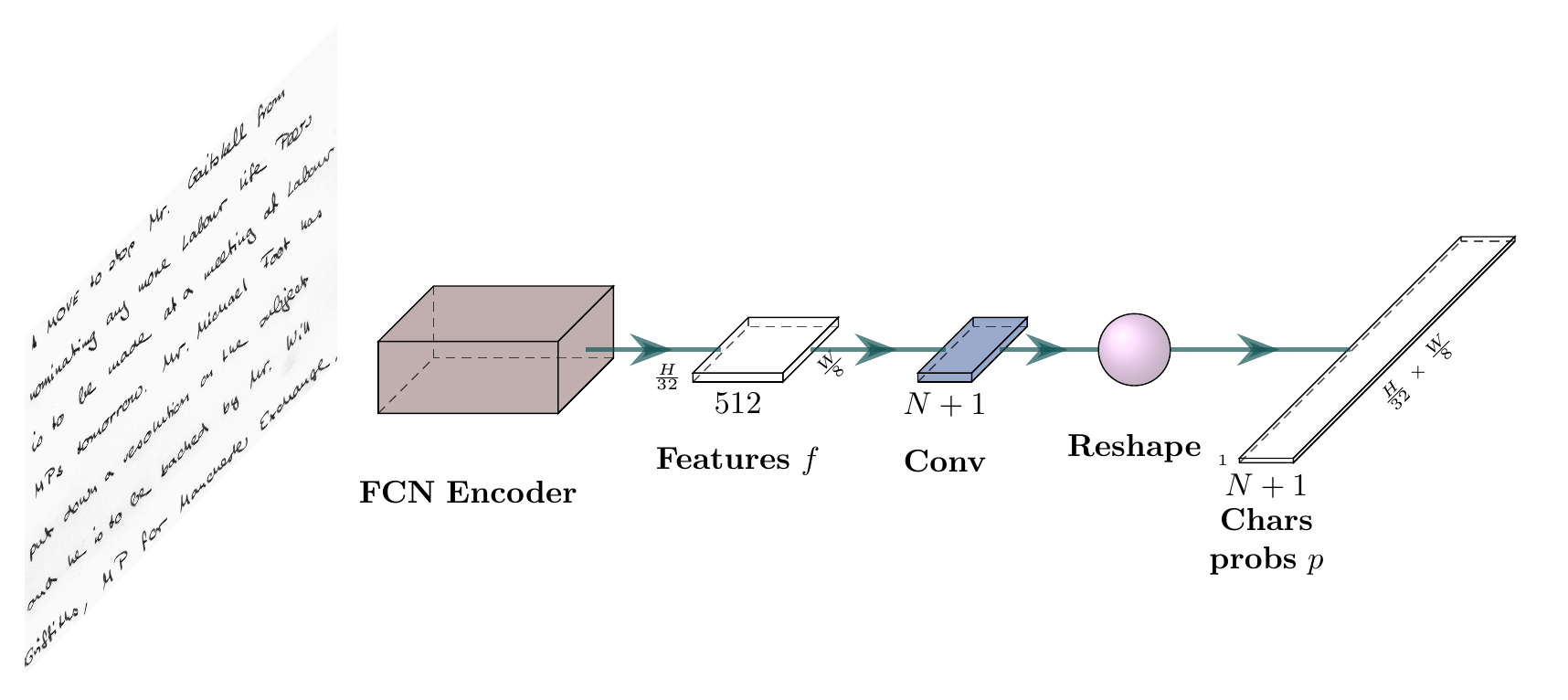}
    \caption{Global architecture overview.}
    \label{fig:archi-overview}
    \end{subfigure}
    
    \begin{subfigure}{\textwidth}
        \centering
        \includegraphics[width=\linewidth]{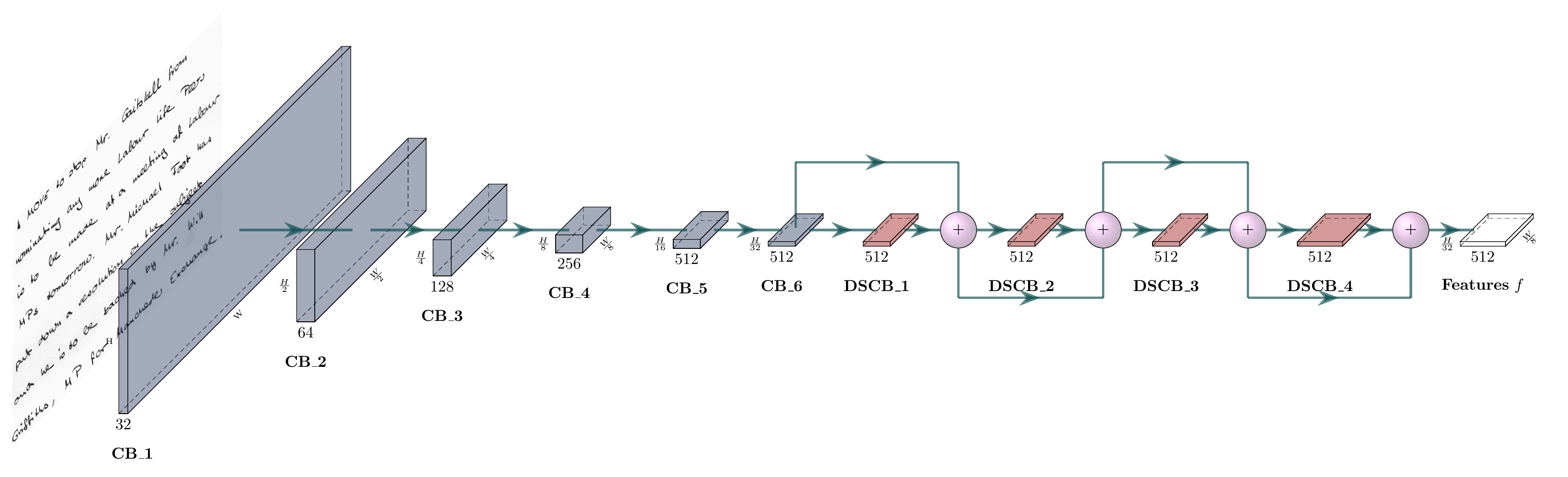}
        \caption{FCN Encoder overview. CB: Convolution Block, DSCB: Depthwise Separable Convolution Block.}
        \label{fig:encoder-overview}
    \end{subfigure}
    
    \caption{Model visualization. \ref{fig:archi-overview} presents an overview of the architecture and \ref{fig:encoder-overview} focuses on the encoder.}
    \label{fig:model-overview}
\end{figure*}

We propose an end-to-end model to perform the optical character recognition of paragraphs. We wanted to keep the original shapes of the input images in order to preserve both their ratio and their details as well as to be flexible enough to adapt to a large variety of datasets. To this end, we use a Fully Convolutional Network as the encoder to analyse the 2D paragraph images. An implicit line segmentation is performed by reducing the vertical axis through row concatenation, reshaping the 2D latent space into a 1D latent space, acting as a collapse operator for this dimension. The training process is based on the standard CTC loss that aligns the label sequence with the data in the 1D latent space without any needs for line breaks in the annotation.
Figure \ref{fig:archi-overview} shows an overview of the model architecture: it consists of a FCN encoder which extracts the features. Then, a convolutional layer predicts the character probabilities. Finally, the rows of predictions are concatenated to obtain one single large row of predictions. This brings us back to a one-dimensional sequence alignment problem which is handled with the standard CTC loss.

    

\subsection{Encoder}
The purpose of the encoder is to extract features from the input images. It implies some convolutions with stride in order to reduce the memory consumption: it takes an input image $\tens{X} \in \mathbb{R}^{H \times W \times C}$ and outputs some feature maps $\tens{f} \in \mathbb{R}^{\frac{H}{32} \times \frac{W}{8} \times 512}$ where H, W and C are respectively the height, the width and the number of channels (C=1 for a grayscale image, C=3 for a RGB image). The encoder architecture is depicted in Figure \ref{fig:encoder-overview}. It corresponds to the encoder proposed in \cite{Coquenet2021}, to which the number of channels has been modified going from 16-256 up to 32-512. It is made up of a succession of Convolution Blocks (CB) and Depthwise Convolution Blocks (DSCB).

CB is defined as two convolutional layers followed by instance normalization and a third convolutional layer. This third convolutional layer has a stride of $1 \times 1$ for $CB\_1$, $2 \times 2$ for $CB\_2$ to $CB\_4$ and $2 \times 1$ for $CB\_5$ to $CB\_6$.

DSCB follows the same structure as CB but the convolutional layers are superseded by Depthwise Separable Convolutions \cite{DepthSepConv} in order to reduce the number of parameters at stake. Moreover, the third DSC has always a stride of $1 \times 1$. This enables to introduce residual connections with element-wise sum operator between the DSCB.

For both blocks, convolutional layers have a $3 \times 3$ kernel, $1 \times 1$ padding and are followed by ReLU activations. In addition, Diffused Mix Dropout (DMD)\cite{Coquenet2021} is used with three potential locations inside each block to reduce overfiting.

\subsection{Decoder}
\begin{figure}[htbp!]
\centering
    \includegraphics[width=\textwidth]{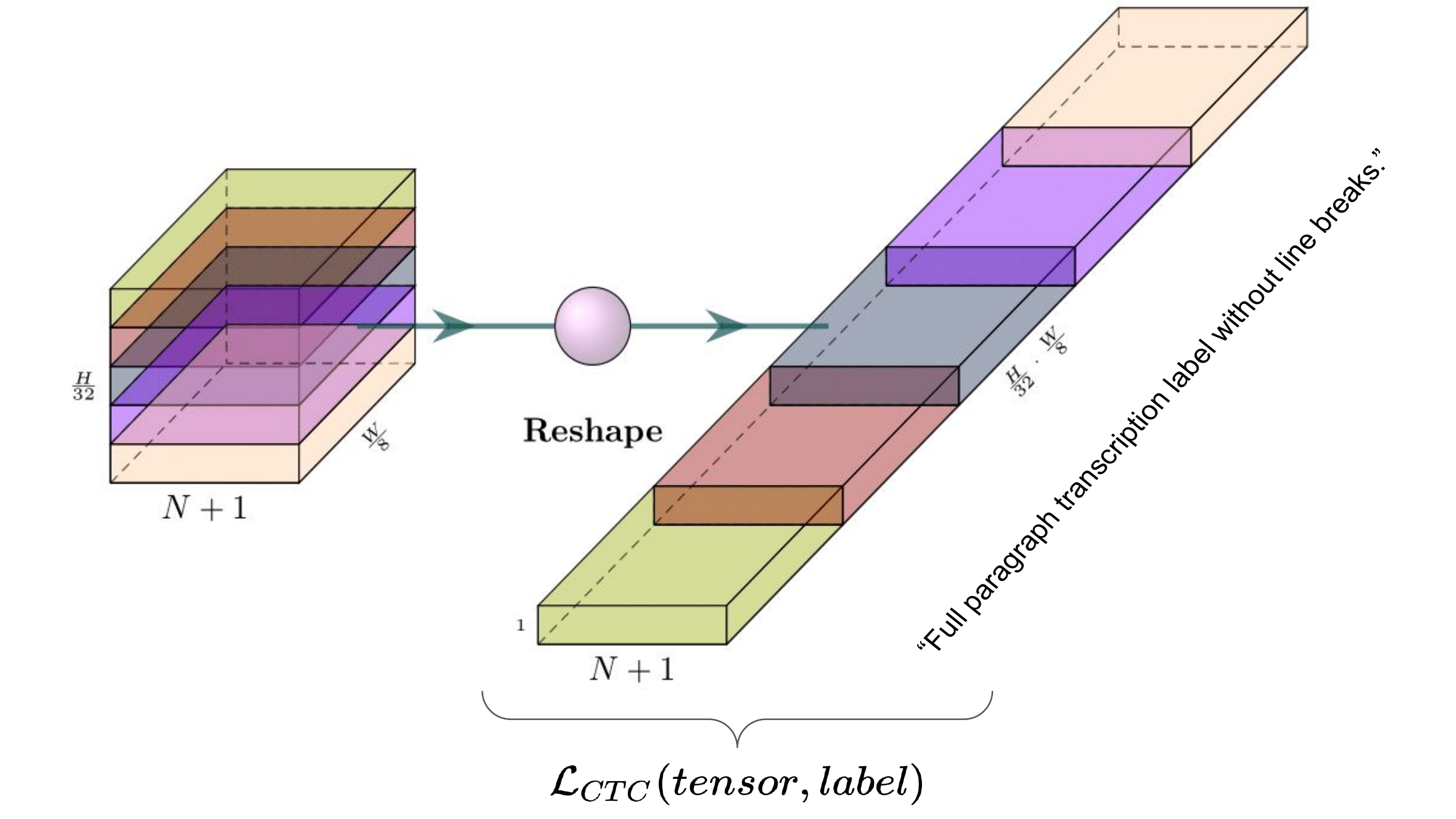}
    \caption{Reshape operation and loss visualization. No computations are performed in the reshape operation, both left and right tensors represent characters and CTC blank label probabilities. The CTC loss is computed between the one-dimensional probabilities sequence and the paragraph transcription.}
    \label{fig:reshape}
\end{figure}

The decoder aims at predicting and aligning the probabilities of the characters and the CTC blank label for each 2D position of the features $\tens{f}$.
The decoder is made up of a single convolutional layer with kernel $5 \times 5$, stride $1 \times 1$ and padding $2 \times 2$. It outputs $N+1$ channels, $N$ being the size of the charset. Finally, the $\frac{H}{32}$ rows are concatenated to obtain the one-dimensional prediction sequence $p\in \mathbb{R}^{(\frac{H}{32} \cdot \frac{W}{8}) \times (N+1)}$ as depicted in Figure \ref{fig:reshape}. The CTC loss is then computed between this one-dimensional prediction sequence and the paragraph transcription ground truth, without line breaks.

We can highlight some important aspects about the decoder:
\begin{itemize}
    \item In this work, the CTC blank label has a new function. Indeed, in standard OCR applied to text lines, the CTC blank label enables to recognize two identical successive characters and to predict "nothing", acting like a joker. Here, it is also used to separate lines in a two-dimensional context, as it allows to label line spacing in the input image.
    \item One should notice that the prediction occurs before reshaping to 1D, which allows to take advantage of the two-dimensional context in the decision layer. This enables to localize the previous and next lines, and to align the predictions of the same text line on the same row \textit{i.e.}, and to separate it from the other text line predictions.
    \item Since the prediction rows are concatenated, they are processed sequentially; nothing prevents the model from predicting the beginning of the text line on one row and the end on the next one as long as there is enough space between this text line and the following one. In Section \ref{section:results}, we show that this allows us to process inclined lines.
\end{itemize}

\section{Experimental study}
\label{experiments}

\subsection{Datasets}
We evaluate our model on three popular datasets at paragraph level: RIMES \cite{RIMES}, IAM \cite{IAM} and READ 2016 \cite{ICFHR_READ2016}.

\subsubsection{RIMES}
We used the RIMES dataset which is made up of French handwritten paragraphs, produced in the context of writing mails scenarios. The images are gray-scaled and have a resolution of 300 dpi. In the official split, 1,500 paragraphs are dedicated to training and 100 paragraphs to evaluation. The last 100 training images are used for validation so as to be comparable with the state of the art. 

\subsubsection{IAM}
The IAM dataset corresponds to handwritten copy of English text passages extracted from the LOB corpus. The images are gray-scaled handwritten paragraph with a resolution of 300 dpi. In this work, we used the unofficial but commonly used split as detailed in Table \ref{table:split}.

\subsubsection{READ 2016}
The READ 2016 dataset corresponds to Early Modern German handwriting. It has been proposed in the ICFHR 2016 competition on handwritten text recognition. It is a subset of the Ratsprotokolle collection, used in the READ project. Images are in color and we used the paragraph level segmentation. We assume that the images have a resolution of around 300 dpi too.

\subsubsection{}
In section \ref{experiments}, some experiments implies pretraining using the line level images of these three datasets. The corresponding splits are shown in Table \ref{table:split}.

\begin{table}[!h]
    \caption{Datasets split in training, validation and test sets and associated charset size}
    \centering
    \resizebox{0.75\linewidth}{!}{
    \begin{tabular}{ c c c c c c}
    \hline
    Dataset & Level & Training & Validation & Test & Charset size\\ 
    \hline
    \hline    
    \multirow{2}{*}{RIMES} & Line & 10,532 & 801 & 778 & \multirow{2}{*}{100}\\
     & Paragraph & 1,400 & 100 & 100 & \\
     
    \hline   
    \multirow{2}{*}{IAM} & Line & 6,482 & 976 & 2,915 & \multirow{2}{*}{79}\\
     & Paragraph & 747 & 116 & 336 & \\
     
    \hline  
    \multirow{2}{*}{READ 2016} & Line & 8,349 & 1,040 & 1,138 & \multirow{2}{*}{89}\\
     & Paragraph & 1,584 & 179 & 197 & \\
     
    \hline
    \end{tabular}
    }
    \label{table:split}
\end{table}

\begin{figure*}[htbp!]
\centering
\begin{minipage}[c]{0.45\textwidth}
    \begin{subfigure}[b]{\textwidth}
    \centering
    \includegraphics[width=\textwidth, frame]{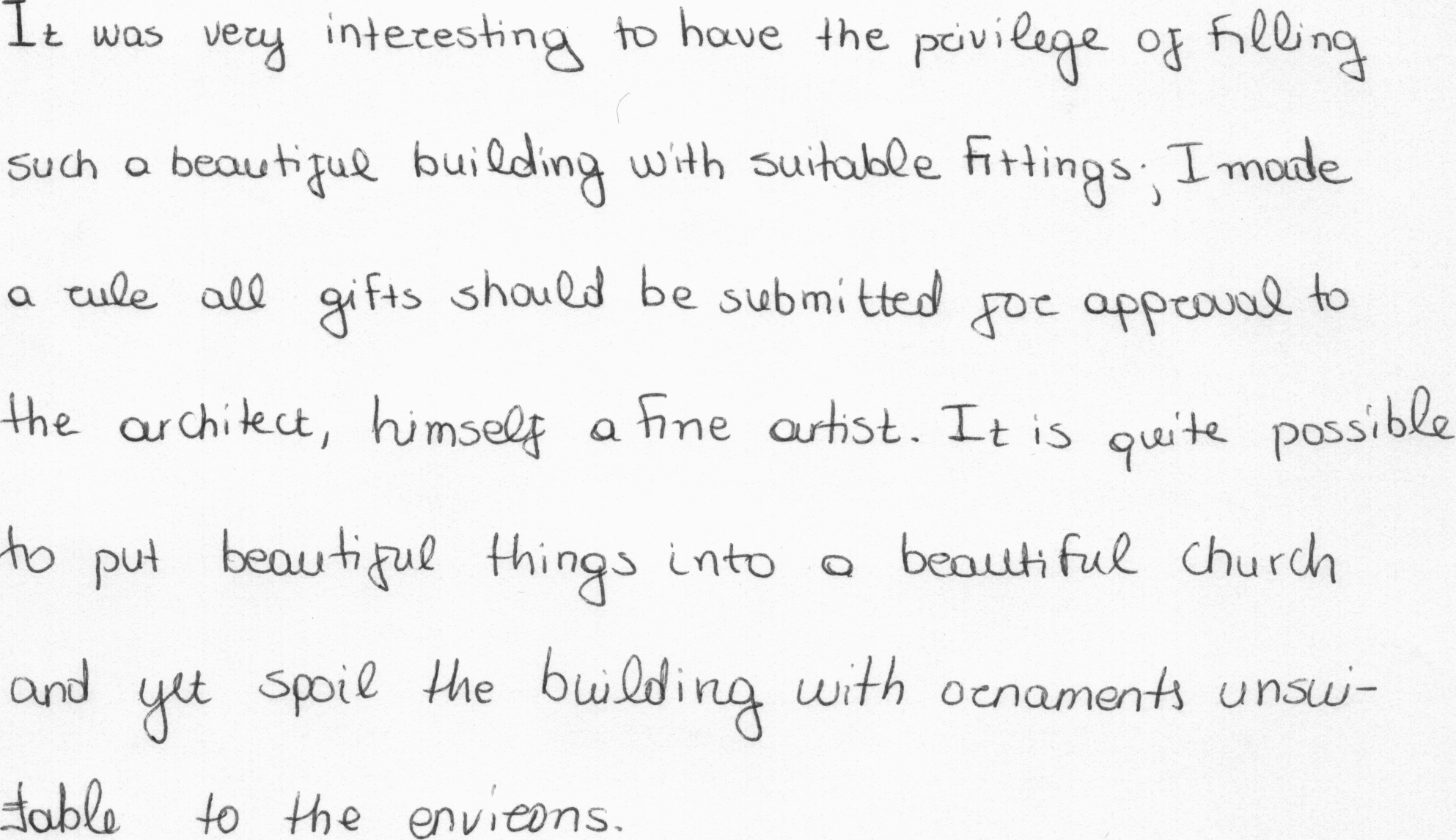}
    \caption{IAM}
    \end{subfigure}
    
    \begin{subfigure}[b]{\textwidth}
    \centering
    \includegraphics[width=\textwidth, frame]{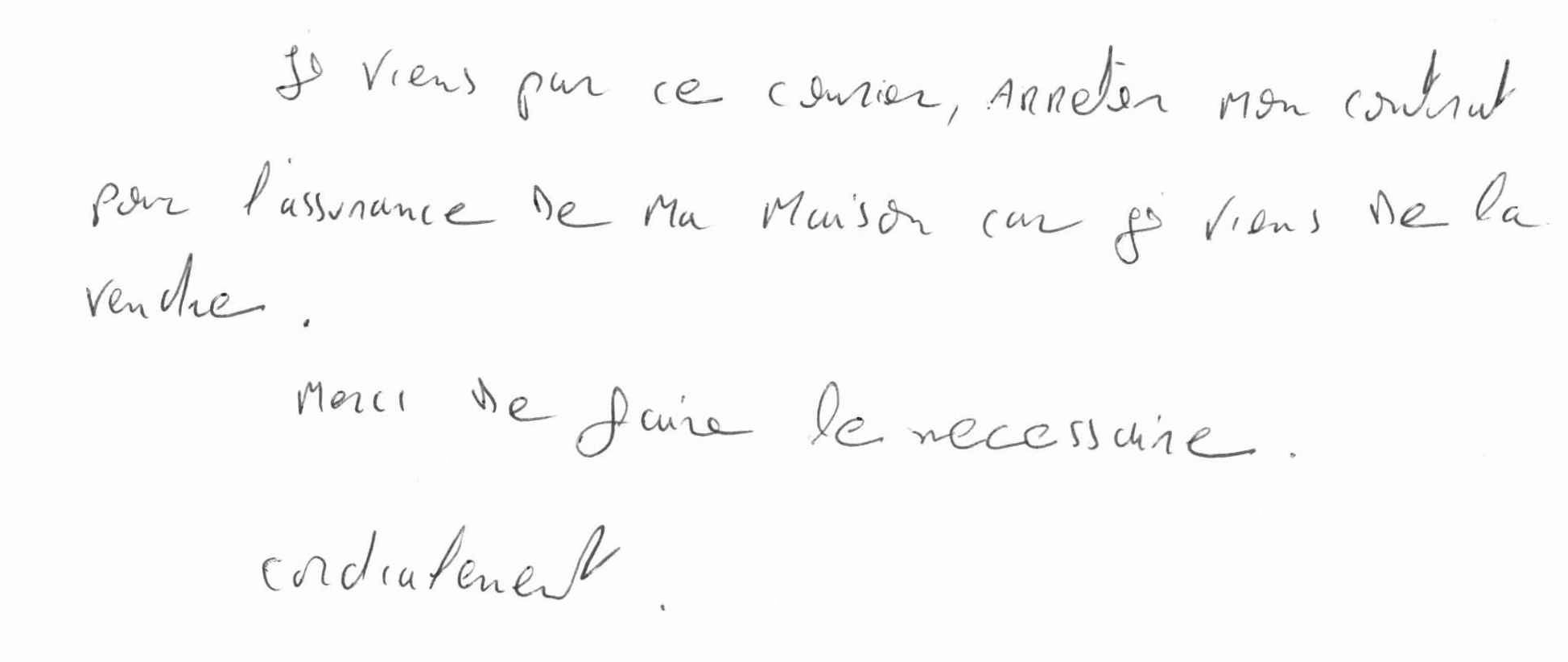}
    \caption{RIMES }
    \end{subfigure}
\end{minipage}
\hfill
\begin{minipage}[c]{0.45\textwidth}
    \begin{subfigure}{\textwidth}
        \centering
        \includegraphics[width=\linewidth, frame]{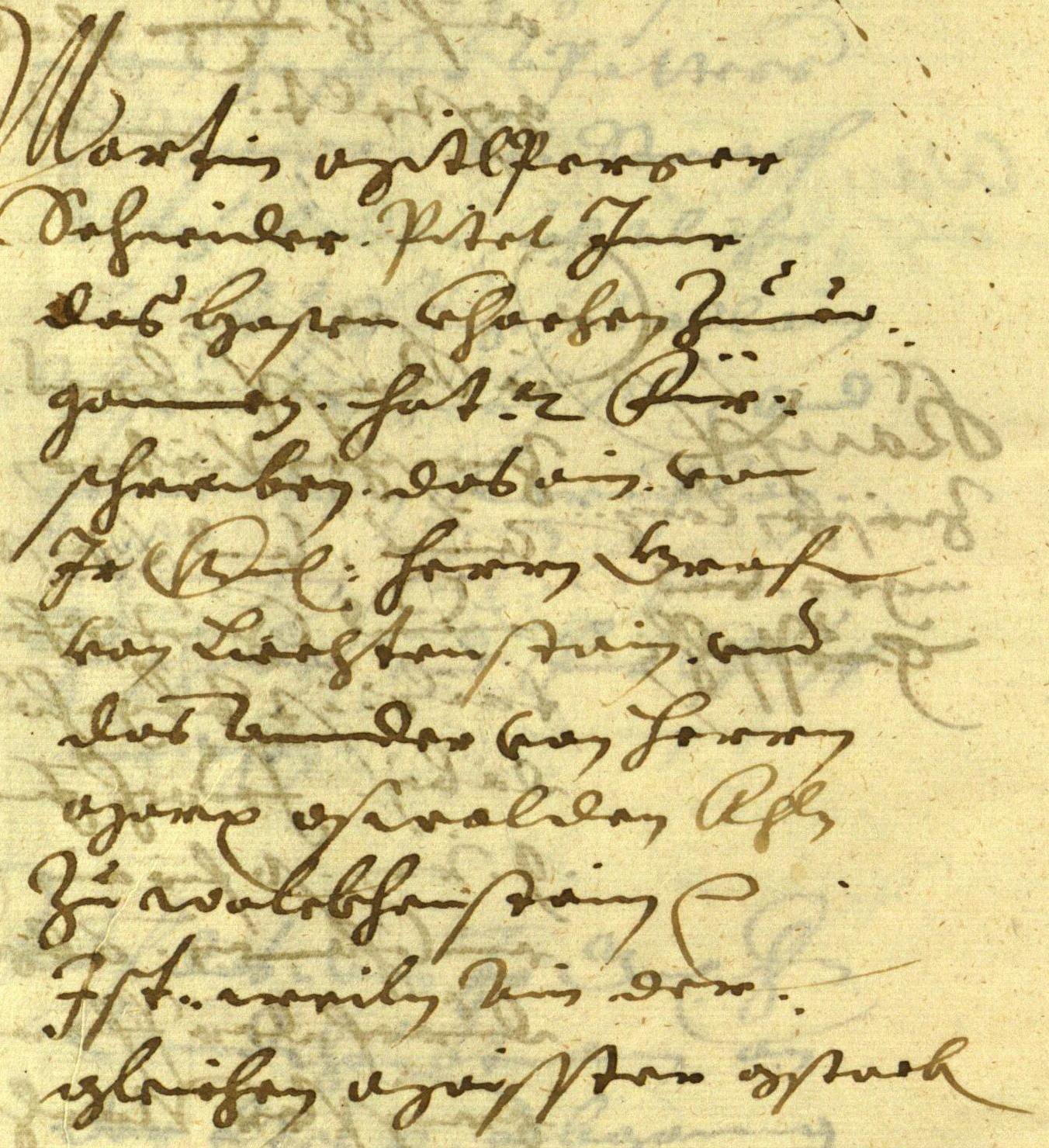}
        \caption{READ 2016}
    \end{subfigure}
\end{minipage}

    \caption{Paragraph image examples from the RIMES, IAM and READ 2016 datasets.}
    \label{fig:dataset}
\end{figure*}

Paragraph image examples from these three datasets are depicted in Figure \ref{fig:dataset}. IAM layout is the more structured and regular. RIMES brings some irregularities in terms of line spacing, text inclination and horizontal text alignment. Finally, the READ 2016 dataset is more complex in terms of noise, text line separation (due to ascents and descents) and size variety. 

\subsection{Preprocessing}
Paragraph images are downscaled by a factor of 2 through a bilinear interpolation leading to a resolution of 150 dpi. Gray-scaled images are converted into RGB images concatenating the same values three times, for transfer learning purposes. They are then normalized (zero mean and unit variance) considering the channels independently.

\subsection{Data Augmentation}
Data augmentation is applied at training time to reduce over-fitting. The augmentation techniques are used in this order: resolution modification, perspective transformation, elastic distortion and random projective transformation (from \cite{Yousef2020}), dilation and erosion, brightness, and contrast adjustment and sign flipping. Each transformation has a probability of 0.2 to be applied. Except for perspective transformation, elastic distortion and random projective transformation which are mutually exclusive, each augmentation technique can be combined with the others.

\subsection{Metrics}
The Character Error Rate (CER) and the Word Error Rate (WER) are used to evaluate the quality of the text recognition. They are both computed with the Levenshtein distance between the ground truth text and the predicted text at paragraph level, without line breaks. Those edit distances are then normalized by the length of the ground truth.
Other metrics are provided in the following experiments such as the number of parameters implied by the models. 

\subsection{Training details}
We used the Pytorch framework to train and evaluate our models. In all experiments, the networks are trained with the Adam optimizer, with an initial learning rate of $10^{-4}$. Trainings are performed on a single GPU Tesla V100 (32Gb), during 2 days, with a mini-batch size of 4 for paragraph images and 16 for text lines images.

\subsection{Additional information}
We do not use any post-processing \textit{i.e.} we do not use any language model nor lexicon constraint. Moreover, we only use best path decoding to get the final predictions from the character probabilities lattice. We use exactly the same training configuration from one dataset to another, without model modification, except for the last layer which depends on the charset size.

\subsection{Results}
\label{section:results}

\subsubsection{Comparison with state of the art}

In this section, we compare our approach to state-of-the-art models on the RIMES, IAM and READ 2016 datasets, at paragraph level and in the same conditions \textit{i.e.} without language model nor lexicon constraint.

Prior to compare the obtained results to the state of the art, it is important to understand the experimental conditions of each of the methods. Table \ref{table:comparison} shows model details that should be taken into account to fairly compare the following tables of results. Quantitative metrics are computed for the IAM dataset, without automatic mixed precision (for a fair comparison with respect to the memory usage). From left to right, the columns respectively denote the architecture, the number of trainable parameters, the maximum GPU memory usage during training (data augmentation included), the minimum transcription level required, the minimum segmentation level required, the use of PreTraining (PT) on subimages, the use of specific Curriculum Learning (CL) and finally the Hyperparameter Adaptation (HA) requirements from one dataset to another.

\begin{table*}[!h]
    \caption{Requirements comparison of the \modelacc{} with the state-of-the-art approaches.}
    \centering
    \resizebox{\linewidth}{!}{
    \begin{tabular}{ l c c c c c c c c}
    \hline
    Architecture & \# Param. & Max memory & Transcription label & Seg. label & PT & CL & HA\\
    \hline
    \hline
    \cite{Carbonell2019} RPN+CNN+BLSTM & & & Word & Word & \\ 
    \cite{Chung2020} RPN+CNN+BLSTM & & & Word & Word & \\
    \cite{Wigington2018} RPN+CNN+BLSTM & & & Line & Line \\
    \cite{Coquenet2021} FCN+LSTM*  & 2.7 M & 2.2 Gb &  Paragraph + line breaks & Paragraph & Line & \xmark & \xmark\\
    \cite{Bluche2017} CNN+MDLSTM** &   & & Paragraph  & Paragraph & Line & \checkmark & \xmark \\
    \cite{Bluche2016} CNN+MDLSTM*  & & & Paragraph & Paragraph & Line & \checkmark & \xmark\\
    \cite{Yousef2020} GFCN & 16.4 M & 8.8 Gb  & Paragraph & Paragraph & \xmark & \xmark & \checkmark\\
    {[}This work - SPAN{]}  FCN &  19.2 M & 5.1 Gb &  Paragraph & Paragraph & Line & \xmark & \xmark\\
    \hline
    * with line-level attention\\
    ** with character-level attention\\
    
    \end{tabular}
    }
    \label{table:comparison}
\end{table*}

As one can see, models from \cite{Carbonell2019,Chung2020,Wigington2018} require transcription and segmentation labels at word or line levels to be trained, which implies more costly annotations. The models from \cite{Bluche2016,Bluche2017,Coquenet2021} and the \modelacc{} are pretrained on text line images to speed up convergence and to reach better results, thus also using line segmentation and transcription labels even if it is not strictly necessary. While the model from \cite{Coquenet2021} needs line breaks in the transcription annotation, \cite{Bluche2016,Bluche2017} used a specific curriculum learning method for training. In \cite{Yousef2020}, some hyperparameters must be modified from one dataset to another in order to reach optimal performance, namely the fixed input dimension and two intermediate upsampling sizes which are crucial. We do not have such problem since we are working with input images of variable size and we focus on the resolution to be robust to the variety of datasets. Moreover, despite a larger number of parameters (+ 17\% compared to \cite{Yousef2020}), the \modelacc{} requires less GPU memory which is a critical point when training deep neural networks.

Table \ref{table:rimes-pg}, \ref{table:iam-pg} and \ref{table:read2016-pg} show the results of the \modelacc{} compared to the state-of-the-art approaches, for the RIMES, IAM and READ 2016 datasets respectively. One can notice that we reach competitive results on those three datasets, each having its own complexities, without any hyperparameter adaptation. Results here includes model pretraining on line images but the model can be trained without pretraining \textit{i.e.} without using any line-level annotation, while keeping competitive results, as shown in Section \ref{section:pretrain}.

\begin{table}[!h]
    \caption{Comparison of the \modelacc{} results with the state-of-the-art approaches at the paragraph level on the RIMES dataset.}
    \centering
    \resizebox{0.7\linewidth}{!}{
    \begin{tabular}{ l c c c c }
    \hline
    \multirow{2}{*}{Architecture} & CER (\%) & WER (\%) & CER (\%) & WER (\%) \\ 
    & validation & validation & test & test\\
    \hline
    \hline
    \cite{Bluche2016}  & 2.5 & 12.0 & 2.9 & 12.6 \\
    \cite{Wigington2018}  & & & 2.1 & 9.3 \\ 
    \cite{Coquenet2021} & \textbf{1.74} & \textbf{8.72} & \textbf{1.90} & \textbf{8.83}  \\
    This work - \modelacc{} & 3.56 & 14.29 & 4.17 & 15.61 \\
    \hline
    \end{tabular}
    }
    \label{table:rimes-pg}
\end{table}

\begin{table*}[!h]
    \caption{Comparison of the \modelacc{} results with the state-of-the-art approaches at the paragraph level on the IAM dataset.}
    \centering
    \resizebox{0.7\linewidth}{!}{
    \begin{tabular}{ l c c c c}
    \hline
    \multirow{2}{*}{Architecture} & CER (\%) & WER (\%) & CER (\%) & WER (\%) \\ 
    & validation & validation & test & test \\
    \hline
    \hline
    \cite{Carbonell2019}*& 13.8 & & 15.6 & \\ 
    \cite{Chung2020} & & & 8.5 & \\
    \cite{Wigington2018}& & & 6.4 & 23.2  \\
    \cite{Coquenet2021}& \textbf{3.04} & \textbf{12.69} & \textbf{4.32} & \textbf{16.24} \\
    \cite{Bluche2017}& &  & 16.2 &  \\
    \cite{Bluche2016}& 4.9 & 17.1 & 7.9 & 24.6 \\
    \cite{Yousef2020} & & & 4.7 & \\
    This work - \modelacc{}  & 3.57 & 15.31 & 5.45 & 19.83 \\
    \hline
    \multicolumn{5}{l}{*Results are given for page level}\\
    \end{tabular}
    }
    \label{table:iam-pg}
\end{table*}

\begin{table}[!h]
    \caption{Comparison of the \modelacc{} results with the state-of-the-art approaches at the paragraph level on the READ 2016 dataset.}
    \centering
    \resizebox{0.7\linewidth}{!}{
    \begin{tabular}{ l c c c c}
    \hline
    \multirow{2}{*}{Architecture} & CER (\%) & WER (\%) & CER (\%) & WER (\%) \\ 
    & validation & validation & test & test\\
    \hline
    \hline
    \cite{Coquenet2021}  & \textbf{3.75} & \textbf{18.61} & \textbf{3.63} & \textbf{16.75} \\
    This work - \modelacc{} & 5.09 & 23.69 & 6.20 & 25.69\\
    \hline
    \end{tabular}
    }
    \label{table:read2016-pg}
\end{table}

\subsubsection{\modelacc{} prediction visualization}

\begin{figure*}[h!]
\begin{minipage}[b]{0.55\textwidth}
    \includegraphics[width=\textwidth, frame]{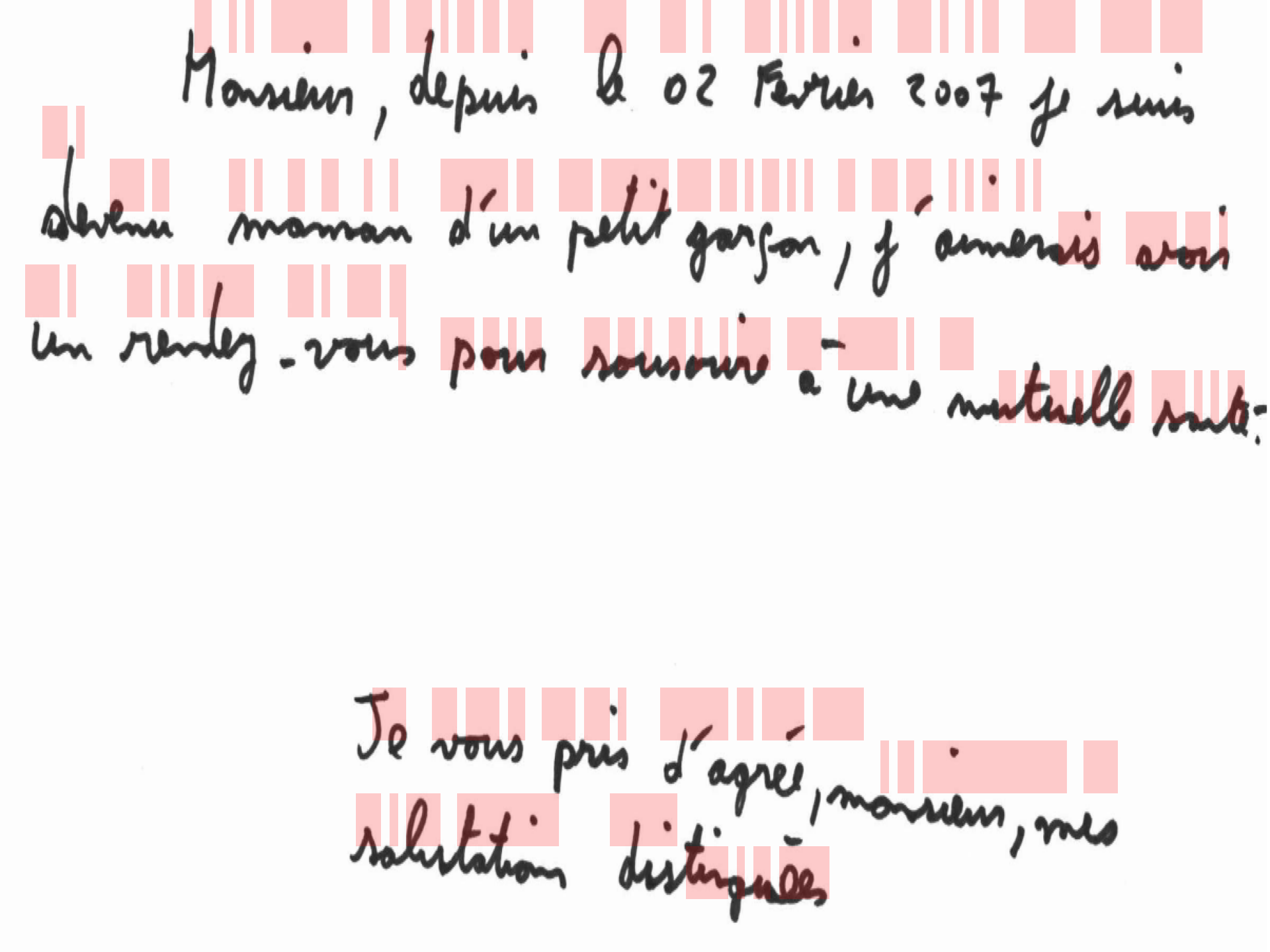}
\end{minipage}
\begin{minipage}[b]{0.45\textwidth}
\scriptsize{
Monsieur, depuis la 02 Ferries 2007 je suis\\
 \\
 de\\
enu maman d'un petit garçon, j'aimer\\
ais avoir\\
 un rendez-vous\\
s pour sousoire à une m\\
ntuelle sont\\
 \\
 \\
 \\
 \\
 \\
Je vous pris d'agrée, m\\
ansieurs, mes\\
 salutation dis\\
tinguées\\
}
\end{minipage}
\par\medskip
Monsieur, depuis l\textbf{a} 02 Fe\textbf{r}rie\textbf{s} 2007 je suis de\textbf{\textit{v}}enu maman d'un petit garçon, j'aimerais avoir un rendez-vous\textbf{s} pour sous\textbf{o}ire à une m\textbf{n}tuelle s\textbf{o}nt\textbf{\textit{é. }}Je vous pris d'agrée, m\textbf{a}nsieurs, mes salutation\textbf{\textit{s}} distinguées

    \caption{\modelacc{} predictions visualization for a RIMES test example. Left: 2D characters predictions are projected on the input image. Red color indicates a character prediction while transparency means blank label prediction. Right: row by row text prediction. Bottom: full text prediction where errors are shown in bold and missing letters are shown in italic.}
    \label{fig:viz}
\end{figure*}

Figure \ref{fig:viz} presents a visualization of the \modelacc{} prediction for an example of the RIMES test set. Character predictions are shown in red; they seem like rectangle since they are resized to fit the input image size (the features size is $\frac{H}{32} \times \frac{W}{8}$). Combined with the receptive field effect, this explains the shift that can occur between the prediction and the text. As one can notice, text line predictions are totally aligned, or aligned by blocks; the lines are well separated by blank labels, which act as line spacing labels. As one can see, this block alignment enables to handle downward inclined lines, especially for lines 3 and 4. Moreover, the model does not degrade in the presence of large line spacing.

    

\subsubsection{Impact of pretraining}
\label{section:pretrain}

In this experiment, we try to highlight the impact of pretraining on the \modelacc{} results. To this end, we compare two pretraining methods at line level: one focusing only on the optical recognition task and the second one focusing on both recognition and prediction alignment. Let's define the following training approaches:
\begin{itemize}
    \item \modelacc{}-Line-R\&A: the \modelacc{} is trained with line-level images. Here, the network has to learn both the recognition and the alignment tasks. 
    \item Pool-Line-R: a new model is trained with line-level images to only focus on the recognition task. This network consists in the previous defined \modelacc{} encoder followed by an Adaptive MaxPooling to collapse the vertical axis; then, a convolutional layer predicts the characters and blank label probabilities. This is the standard way to process text line images, as in \cite{Coquenet2020}. Since the prediction is already in one dimension, the network does not need to care about vertical alignment.
    \item \modelacc{}-Scratch: the \modelacc{} is trained directly on paragraph images without pretraining.
    \item \modelacc{}-PT-R: \modelacc{} weights are initialized with Pool-Line-R ones. It is then trained with paragraph images.
    \item \modelacc{}-PT-R\&A: \modelacc{} weights are initialized with \modelacc{}-Line-R\&A ones. It is then trained with paragraph images.
\end{itemize}

One can note that the vertical receptive field is bigger than line image heights. Thus, when the model switches from line images to paragraph images, the decision benefits from more context, which replaces part of the previously used padding.

Results are given in Table \ref{table:pretrain}. Focusing on the line-level section, one can notice that, as expected, we reached better results on text lines when the task is reduced to optical recognition compared to the task of recognition and alignment, whatever the dataset. This leads to a CER improvement of 0.94 point for IAM, 0.79 point for RIMES and 0.38 point for READ 2016. 
Now, comparing the paragraph level approaches, one can notice that, except for the RIMES CER, pretraining leads to better results, and sometimes by far (-2.93 points of CER for READ 2016); moreover, pretraining on an easier task, \textit{i.e.} only on the optical recognition, is even more efficient.

\begin{table}[!h]
    \caption{Impact of pretraining the SPAN on line images for the IAM, RIMES and READ 2016 datasets. Results are given on the test sets.}
    \centering
    \resizebox{\linewidth}{!}{
    \begin{tabular}{ l c c c c c c}
    \hline
    \multirow{2}{*}{Approach}& \multicolumn{2}{c}{IAM} & \multicolumn{2}{c}{RIMES} & \multicolumn{2}{c}{READ 2016} \\
     & CER (\%) & WER (\%) & CER (\%) & WER (\%) & CER (\%) & WER (\%) \\ 
    \hline
    \hline
    \textbf{Line-level training} \\
    Pool-Line-R & \textbf{4.82} & \textbf{18.17} & \textbf{3.02} & \textbf{10.73} & \textbf{4.56} & \textbf{21.07}\\
    \modelacc{}-Line-R\&A & 5.76 & 21.33 & 3.81 & 13.80 & 4.94 & 22.19\\   
    \\
    \textbf{Paragraph-level training} \\
    \modelacc{}-Scratch & 6.46 & 23.75 & \textbf{4.15} & 16.31 & 9.13 & 36.63\\
    \modelacc{}-PT-R & \textbf{5.45} & \textbf{19.83} & 4.74 & \textbf{15.55} & \textbf{6.20} & \textbf{25.69}\\
    \modelacc{}-PT-R\&A & 5.78 & 21.16 & 4.17 & 15.71 & 6.62 & 27.38\\
    \hline
    \end{tabular}
    }
    \label{table:pretrain}
\end{table}

\begin{figure*}[h!]
\centering
\begin{minipage}[c]{0.45\textwidth}
\begin{subfigure}[b]{\textwidth}
    \centering
    \includegraphics[width=\textwidth]{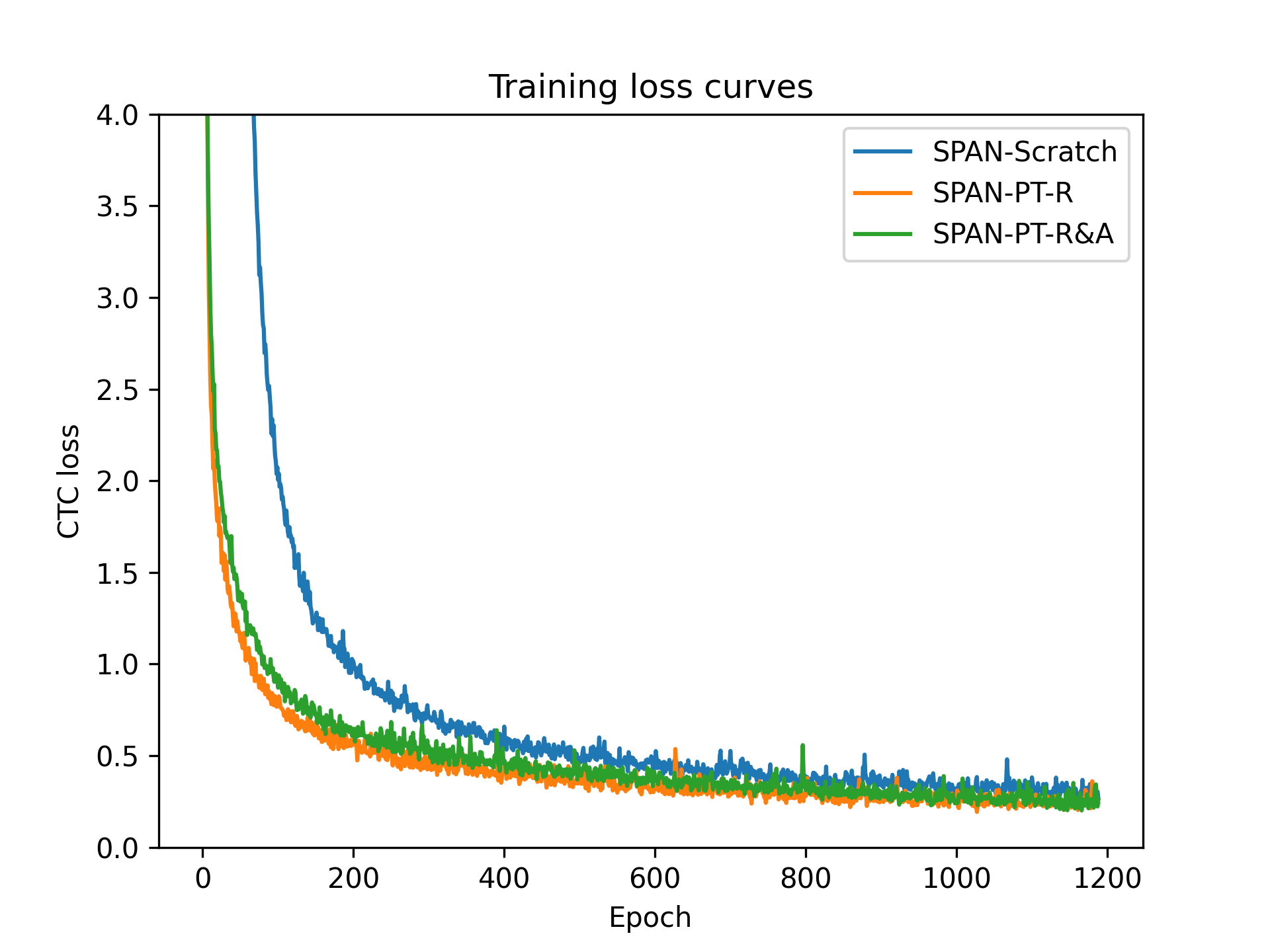}
    \caption{IAM}
    \end{subfigure}
\end{minipage}
\hfill
\begin{minipage}[c]{0.45\textwidth}
    \begin{subfigure}{\textwidth}
        \centering
        \includegraphics[width=\linewidth]{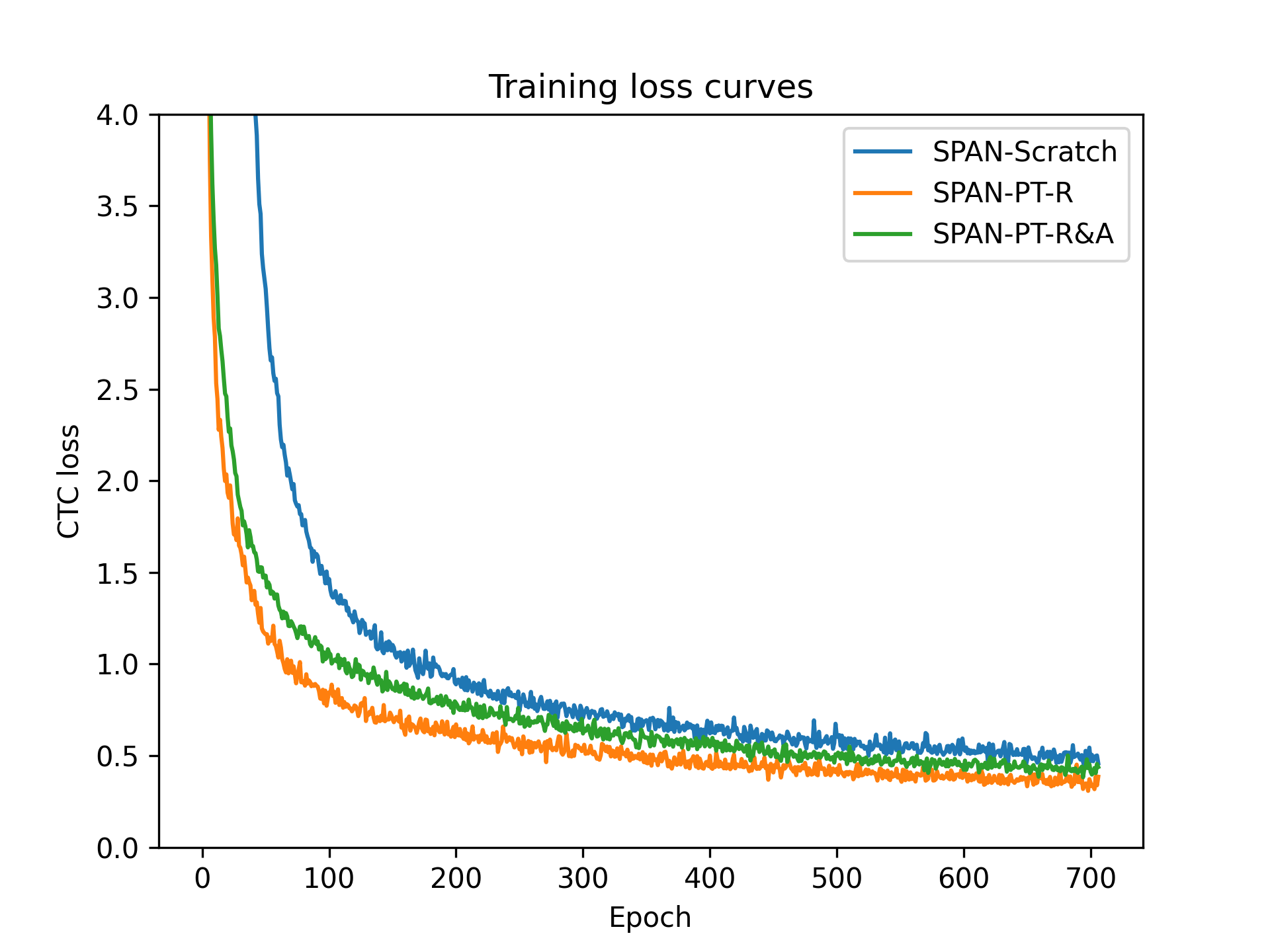}
        \caption{RIMES}
    \end{subfigure}
\end{minipage}

\begin{minipage}[c]{0.45\textwidth}
    \begin{subfigure}{\textwidth}
        \centering
        \includegraphics[width=\linewidth]{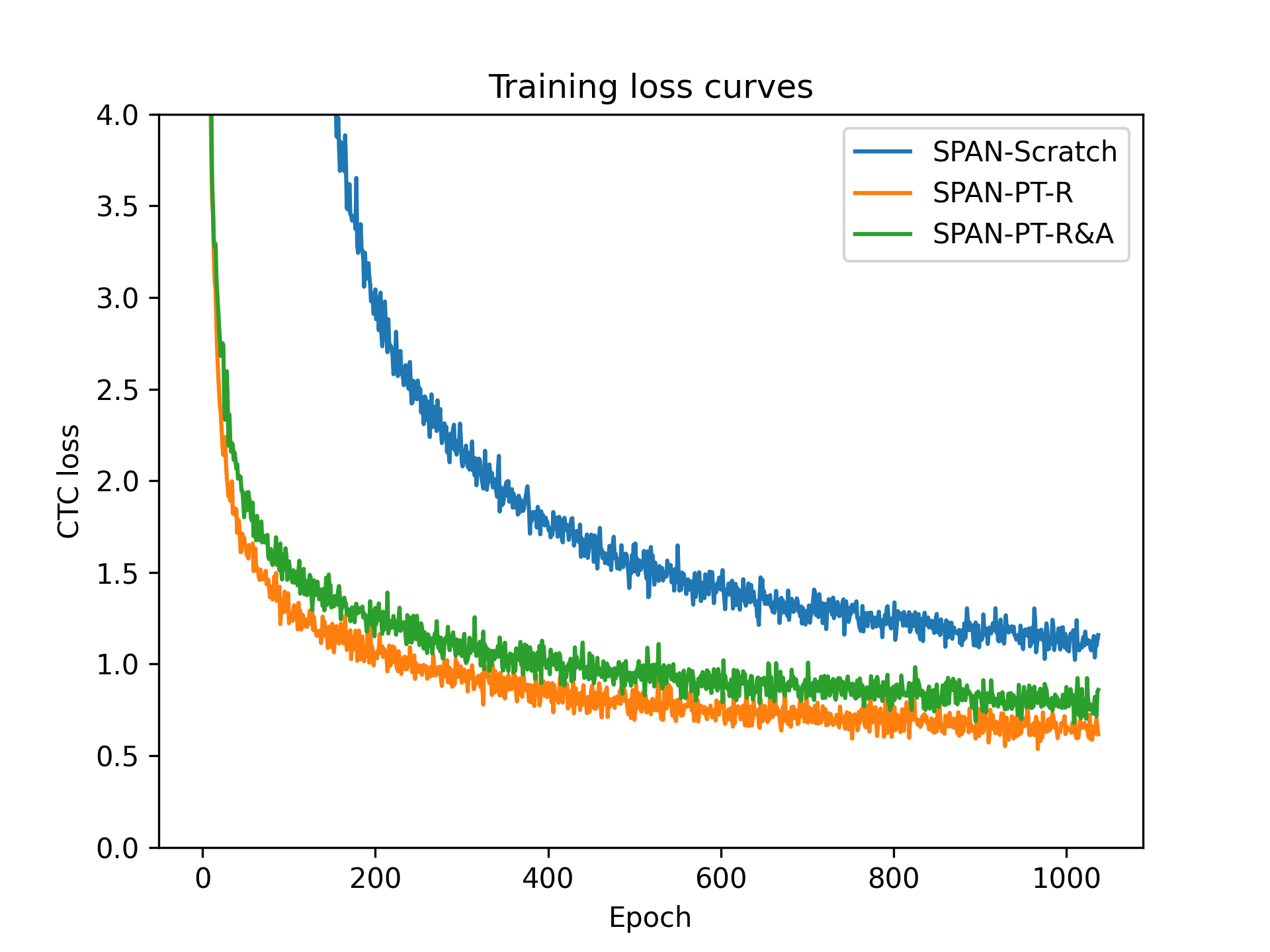}
        \caption{READ 2016}
    \end{subfigure}
\end{minipage}
    
    \caption{Training curves comparison between the different pretraining approaches, on the RIMES, IAM and READ 2016 datasets.}
    \label{fig:losses}
\end{figure*}

The RIMES CER value can be explained by the difference between the CTC loss and the levenshtein distance which are not the same. As a matter of fact, generally, a lower CTC loss implies a lower CER but it is not always true. Indeed, Figure \ref{fig:losses} shows the different loss CTC training curves for the three datasets. This time, we can clearly see that, even for RIMES, pretraining on the recognition task only is more beneficial. This figure also demonstrates the convergence speed up brought by these pretraining approaches.

\subsubsection{Discussion}
As we have seen in Figure \ref{fig:viz}, the 2D prediction keeps the spatial information. As such, we can assume that the \modelacc{} could be used as a primary stage of a deeper end-to-end network that could handle more complex tasks such as word spotting in handwritten digitized document. Moreover, since we are using the standard CTC loss, one can easily add standard character or word language model to further improve the results. However, it has to be noticed that this model is limited to single-column multi-line text images due to the row concatenation operation. Moreover, the \modelacc{} can handle easily handle downward sloping lines but cannot handle upward ones due to the fixed reshaping order.

\section{Conclusion}
\label{conclusion}

In this paper, we proposed the \modelname{}, an end-to-end recurrence-free segmentation-free FCN model performing OCR at paragraph level. It reaches competitive results on the RIMES, IAM and READ 2016 datasets without any model architecture or training adaptation from one to another. It follows a new training approach bringing several other advantages. First, it only needs transcription label at paragraph level (without line breaks), leveraging the need for handmade annotation, which is a critical point for a deep learning system. Second, training this model is as simple as training a line-level OCR with the CTC loss. Finally, it can handle variable image input sizes, making it robust enough to adapt to multiple datasets; it is also able to deal with downward inclined text lines.

\section*{Acknowledgments}
The present work was performed using computing resources of CRIANN (Normandy, France) and HPC resources from GENCI-IDRIS (Grant 2020-AD011012155). This work was financially supported by the French Defense Innovation Agency and by the Normandy region.

\vspace{-0.4cm}
\begin{figure}[!h]
\centering
    \includegraphics[height=1.8cm]{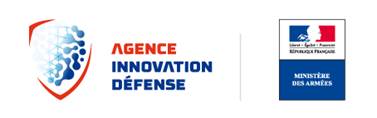}
    ~
    \includegraphics[height=1.8cm]{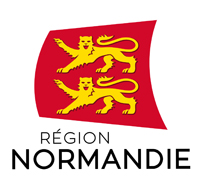}
    \label{data}
\end{figure}
\vspace{-0.5cm}

\bibliographystyle{splncs04}
\bibliography{references.bib}

\end{document}